\documentclass[sigconf,nonacm]{acmart}
\graphicspath{{Figures/}}
\usepackage[
  type={CC},           
  modifier={by-nc-sa}, 
  version={4.0}        
]{doclicense}
\usepackage{float}
\usepackage[T1]{fontenc}    
\usepackage{hyperref}       
\usepackage{newunicodechar}
\usepackage{textcomp} 
\DeclareUnicodeCharacter{2194}{\leftrightarrow}
\newunicodechar{≈}{\approx}
\usepackage{enumitem} 
\newunicodechar{ }{\,} 
\newunicodechar{̈}{\"{}} 
\usepackage{url}            
\usepackage{booktabs}       
\usepackage{multirow}       
\usepackage{amsfonts}       

\usepackage{amssymb}        
\usepackage{amsmath}
\DeclareMathOperator{\cosine}{cosine}
\usepackage{graphicx}        
\usepackage{nicefrac}       
\usepackage{microtype}      
\usepackage{textcomp}       
\usepackage{xcolor}         
\usepackage{graphicx}
\usepackage{tikz}
\usepackage{pgfplots}
\pgfplotsset{compat=1.18}
\usetikzlibrary{shapes.geometric, arrows.meta, positioning, backgrounds, shapes, arrows}
\usepackage{algorithm}
\usepackage{algpseudocode}
\usepackage{amsmath}
\usepackage{enumitem}
\usepackage{caption}
\usepackage{wrapfig}
\usepackage{multirow}
\usepackage{tcolorbox}
\usepackage{multicol}
\usepackage{lipsum}
\usepackage{paracol}
\usepackage{colortbl}  
\usepackage{array} 
\usepackage{colortbl}    
\usepackage{caption}    
\usepackage{float}  
\usepackage{subcaption}
\usepackage{adjustbox}
\usepackage{graphicx}
\usepackage{tikz}
\usepackage{graphicx}
\usepackage{lipsum}
\usetikzlibrary{shapes.geometric, arrows.meta, positioning}
\setcopyright{none}
\tikzstyle{process} = [rectangle, rounded corners, minimum width=2.8cm, minimum height=0.9cm, text centered, draw=black, font=\footnotesize, fill=blue!10]
\tikzstyle{decision} = [diamond, aspect=2, text centered, draw=black, font=\footnotesize, fill=yellow!20]
\tikzstyle{arrow} = [thick,->,>=Stealth]
\tikzstyle{startstop} = [ellipse, minimum width=2.8cm, minimum height=0.9cm, text centered, draw=black, font=\footnotesize, fill=red!10]

\definecolor{darkgreen}{RGB}{0,100,0}

\AtBeginDocument{%
  }
\setcopyright{none}
\DeclareUnicodeCharacter{202F}{\,}

\begin{document}

\title{MultiFinRAG: An Optimized Multimodal Retrieval-Augmented Generation (RAG) Framework for Financial Question Answering}
\subtitle{Preprint Copy}

\author{Chinmay Gondhalekar}
\affiliation{%
  \institution{S\&P Global Ratings}
  \city{New York}
  \country{USA}
}
\email{chinmay.gondhalekar@spglobal.com}

\author{Urjitkumar Patel}
\affiliation{%
  \institution{S\&P Global Ratings}
  \city{New York}
  \country{USA}}
\email{urjitkumar.patel@spglobal.com}

\author{Fang-Chun Yeh}
\affiliation{%
  \institution{S\&P Global Ratings}
  \city{New York}
  \country{USA}
}
\email{jessie.yeh@spglobal.com}


\begin{abstract}

Financial documents—such as 10-Ks, 10-Qs, and investor presentations—span hundreds of pages and combine diverse modalities, including dense narrative text, structured tables, and complex figures. Answering questions over such content often requires joint reasoning across modalities, which strains traditional large language models (LLMs) and retrieval-augmented generation (RAG)\cite{lewis2020rag} pipelines due to token limitations, layout loss, and fragmented cross-modal context. We introduce \emph{MultiFinRAG}, a retrieval-augmented generation framework purpose-built for financial QA. MultiFinRAG first performs \emph{multimodal extraction} by grouping table and figure images into batches and sending them to a lightweight, quantized open-source multimodal LLM, which produces both structured JSON outputs and concise textual summaries.  These outputs, along with narrative text, are embedded and indexed with modality-aware similarity thresholds for precise retrieval. A \emph{tiered fallback strategy} then dynamically escalates from text-only to text+table+image contexts when necessary, enabling cross-modal reasoning while reducing irrelevant context. Despite running on commodity hardware, MultiFinRAG achieves 19 percentage points higher accuracy than ChatGPT-4o\cite{gpt4o} (free-tier) on complex financial QA tasks involving text, tables, images, and combined multimodal reasoning.

\end{abstract}

\keywords{Retrieval-Augmented Generation (RAG), Multimodal Inference, Large Language Models, Natural Language Processing, Financial QA, PDF Document Understanding, Deep Learning}
\maketitle

\section{Introduction}
Modern financial filings often span over a hundred pages and integrate dense narrative text, structured tables, and complex graphical elements. For instance, a recent Morgan Stanley 10-Q contains approximately 120 pages, more than 275 tables, and nearly 200 figures. Accurate question answering (QA) on such filings is critical for analysts, auditors, and automated financial agents involved in risk monitoring, compliance, and investment decision-making. However, addressing queries over these documents poses significant challenges for LLMs and conventional retrieval-augmented generation (RAG) pipelines, with two core issues:
\par\noindent\textbf{Length and cost:} Document length far exceeds LLM token limits, driving up API costs and making end-to-end processing infeasible.
\par\noindent\textbf{Mixed formats:} Structured tables and visual figures lose their inherent relationships when naively flattened into plain text, obscuring the numerical context crucial for financial QA.
Retrieval-augmented generation (RAG) mitigates the length issue by retrieving only relevant passages, but standard RAG pipelines typically:
\begin{itemize}[leftmargin=8pt]
  \item Use fixed-size, non-overlapping text chunks, often fragmenting coherent explanations or splitting numeric context across boundaries.
  \item Treat tables and charts as unstructured text, sacrificing tabular relationships and visual insights.
  \item Employ static top-$k$ retrieval, which can return redundant or marginally relevant snippets and dilute answer quality.
\end{itemize}
We introduce \emph{MultiFinRAG}, a RAG framework tailored for financial QA with three key advances:
\begin{itemize}[leftmargin=8pt]
  \item \textbf{Batch multimodal extraction} : Small groups of table and chart images are fed to a lightweight multimodal LLM, which returns structured JSON plus concise summaries, preserving numerical relationships and visual insights for indexing.
  \item \textbf{Semantic chunk merging \& thresholded retrieval} : Over-segmented text chunks are recombined based on embedding similarity and indexed in FAISS\cite{douze2024faiss} with modality-specific thresholds (e.g.\ 80\% for text, 65\% for images) to filter marginal contexts.
  \item \textbf{Tiered fallback strategy} : Queries first leverage high-similarity text; if insufficient hits remain (below a preset count or similarity), retrieval automatically escalates to table and image context and combines text + table + image context, ensuring comprehensive coverage.
\end{itemize}

\begin{figure}[t]
  \centering
  \fbox{
  \includegraphics[width=\linewidth]{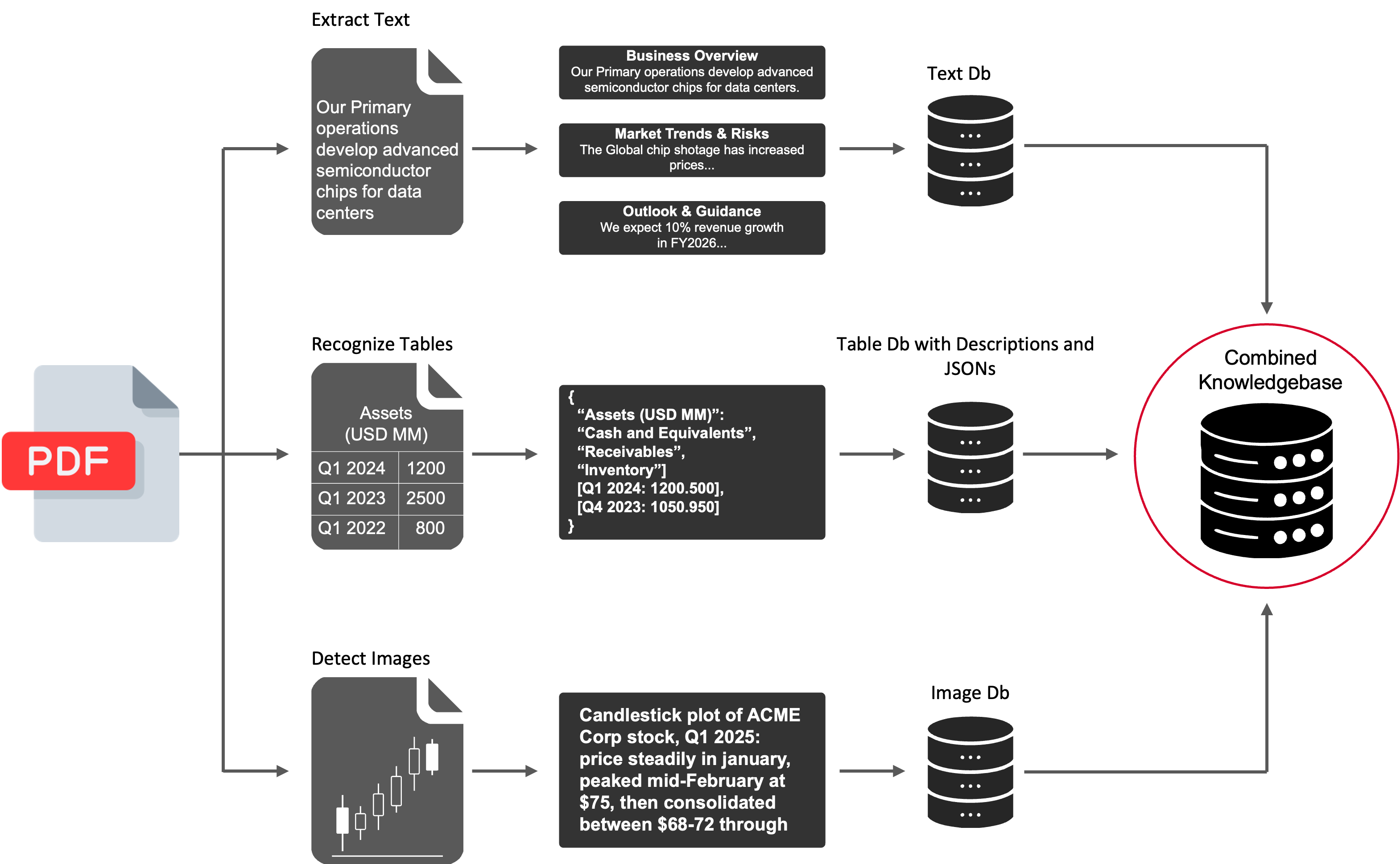}
  \Description{Flow of the MultiFinRAG pipeline.}
  }
  \caption{MultiFinRAG pipeline: knowledge base construction from PDF text, tables, and figures}
  \label{fig:overview}
\end{figure}

\section{Related Work}
\sloppy
The integration of LLMs into real-world decision-making pipelines has propelled the development of Retrieval-Augmented Generation (RAG) systems to address knowledge limitations of LLMs, especially in domain-specific or constantly evolving contexts. Financial documents, particularly regulatory filings like 10-Ks and 8-Ks, present unique challenges for RAG: their content is lengthy, multimodal (text, tables, charts), and often semantically distributed across sections. Existing RAG systems often fail to answer questions that require synthesizing information from multiple modalities or sources.

Open-Source Multimodal Capabilities
Recent advancements in open-source multimodal models—such as Meta's Llama-3.2-11B-Vision-Instruct \cite{touvron2023llama}, Google’s Gemma \cite{team2024gemma}, and DeepSeek \cite{liu2024deepseek}-have made it feasible to build robust, multimodal RAG pipelines without relying on proprietary tools. RAG has matured significantly from its early design as a simple retrieval-augmented QA pipeline. Modern frameworks have introduced dynamic retrieval strategies, smarter chunking mechanisms, and hierarchical or adaptive organization of retrieved content.

Approaches like SELF-RAG \cite{selfrag} empower LLMs to assess and refine their own retrievals through self-reflection. T-RAG \cite{trag} organizes retrieval hierarchically through tree-based entity structures, while MoG (Mix-of-Granularity) \cite{zhong2024mog} and DRAGIN \cite{su2024dragin} dynamically control chunk sizes and retrieval timing based on query characteristics and generation behavior. Late Chunking \cite{gunther2024latechunking} further enhances retrieval alignment by deferring chunk segmentation until after document embedding.

Dense Passage Retrieval (DPR) \cite{dsp}, built on a dual-encoder architecture, remains a cornerstone in RAG systems for open-domain QA, outperforming sparse retrieval baselines. Evaluation frameworks like eRAG \cite{erag} and DPA-RAG \cite{dpa} assess retrieval relevance in the context of downstream generation, while ClashEval \cite{wu2024clasheval} and vRAG-Eval\cite {answer_quality} benchmark LLM performance under conflicting or noisy retrievals.

PDFTriage \cite{saadfalcon2023pdftriagequestionansweringlong} shows that representing structured documents like PDFs as plain text leads to loss of layout and context, and proposes layout-aware retrieval to improve QA over figures, tables, and multi-page content. In finance, Smith et al. \cite{financial_chunking} showed that structural segmentation improves the accuracy of retrieval, while Fin-RAG \cite{kannammal2025fin} improves the accuracy by using tree-based retrieval methods with meta-data clustering.

Despite these advances, current financial RAG systems remain limited in their ability to handle questions that require coordinated reasoning across multiple modalities—text, tables, and figures. Most approaches retrieve relevant snippets but lack mechanisms for aligning and synthesizing information across formats.

To bridge the gap in answering complex, multimodal financial questions, we introduce MultiFinRAG - designed for integrated reasoning across text, tables, and figures. MultiFinRAG combines approximate nearest-neighbor retrieval, modality-aware similarity filtering, and a tiered fallback strategy to handle diverse query contexts with precision. Unlike prior RAG systems, it is specifically tailored for financial documents, emphasizing empirical fidelity and structured reasoning. By preserving alignment between textual, numerical, and visual information in long, layout-rich PDFs, MultiFinRAG addresses a key limitation in existing retrieval-augmented approaches.

\begin{figure}[t]
  \centering
  \fbox{%
    \includegraphics[width=\linewidth]{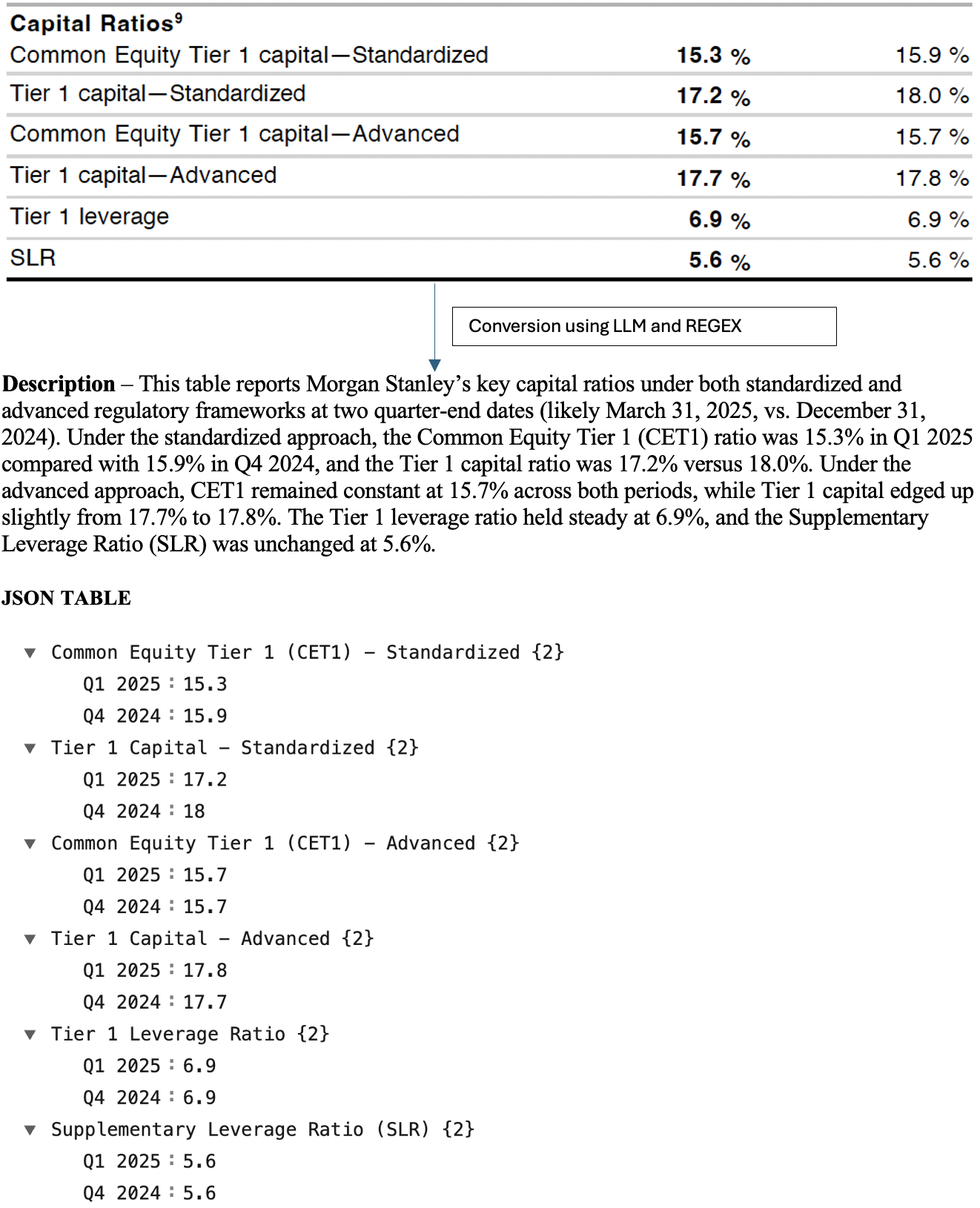}%
  }
  \caption{Generated table description and JSON}
  \label{table_desc}
  \Description{Generated table description and JSON}
\end{figure}


\begin{figure}[b]
  \centering
  \fbox{%
  \includegraphics[width=0.9\linewidth]{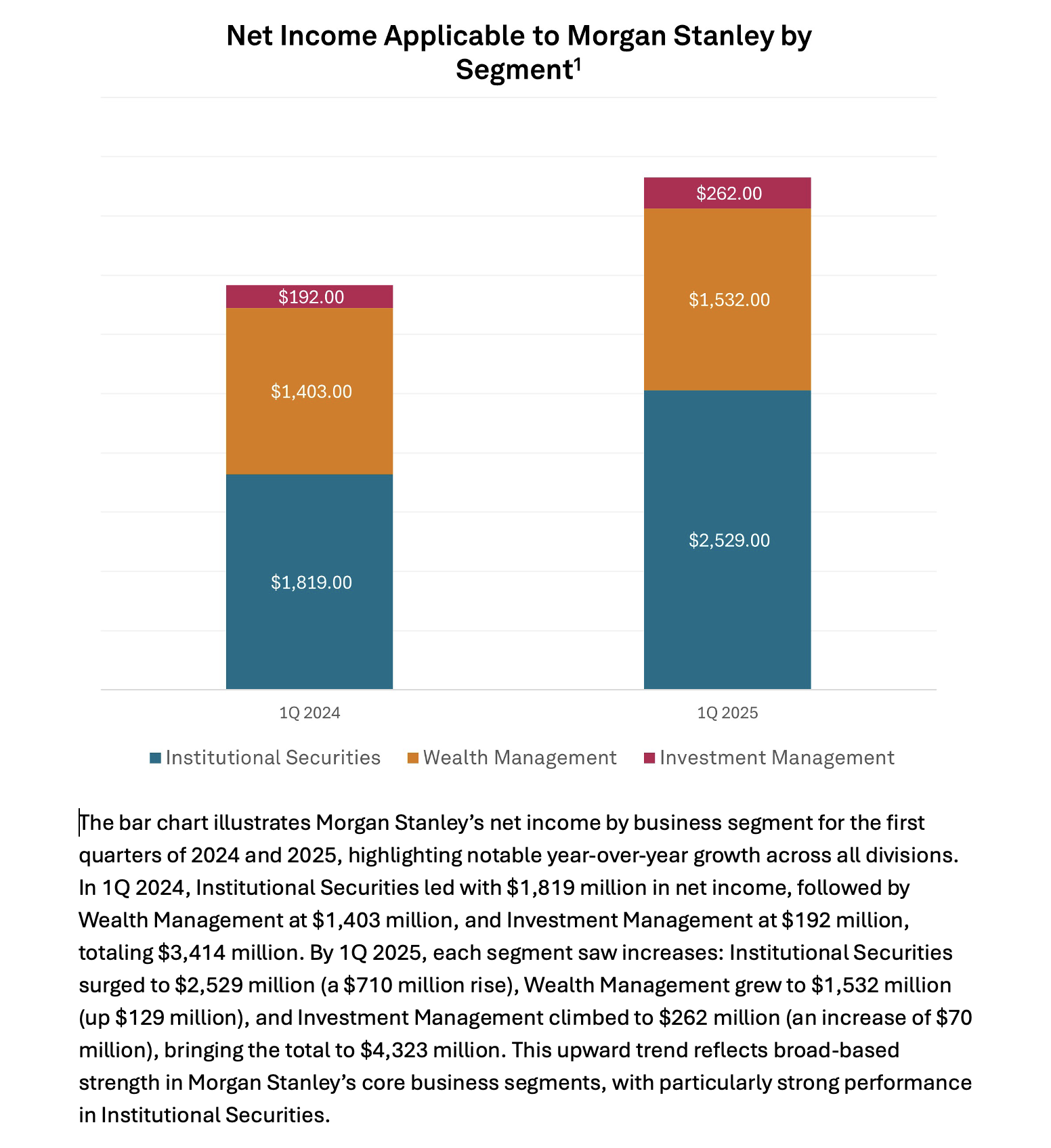}
  }
  \caption{Image Summary Generation Flowchart}
  \label{imagedesc}
  \Description{Image Summary Example}
\end{figure}

\section{Methodology}

\subsection{Models and Tools Used}\label{sec:models-tools}
\sloppy
We utilize a combination of specialized and general-purpose models to handle the multimodal nature of financial documents:
\begin{itemize}[nosep, leftmargin=*]
\item \textbf{Table Detection:} Detectron2Layout \cite{detectron2},
pre-trained on the TableBank\cite{li2020tablebankbenchmarkdatasettable} dataset, identifies and extracts tabular structures.
\item \textbf{Image Detection:} Pdfminer's\cite{pdfminer} layout analysis (\texttt{LTImage/LTFigure}) locates charts and diagrams within PDFs.
\item \textbf{Multimodal Summarization:} Quantized Gemma3: 12B \cite{team2024gemma} and LLama 3.2:11B \cite{llama3.1}vision models were tried for batch summarization of tables and images into structured JSON and plain text summaries, as these are state of the art open source multi-modal capability models.
\item \textbf{Embedding Generation:} BAAI/bge-base-en-v1.5 \cite{baai-base} model using SentenceTransformer generates embeddings for semantic text chunks, tables, and image summaries.
\item \textbf{Approximate Retrieval:} FAISS (IVF-PQ index) provides scalable, efficient approximate nearest-neighbor searches.
\item \textbf{LLMs Used:}
  We integrate two quantized open‐source multimodal LLMs via the Ollama framework, chosen for their strong fusion of text, tables, and images under constrained hardware:
  \begin{itemize}[nosep,leftmargin=*,label=\textbullet]
    \item \emph{Gemma3:12B} \cite{team2024gemma}: demonstrates best‐in‐class multimodal reasoning across narrative, tabular, and visual inputs. Quantization reduces its 24 GB GPU footprint by $\approx$ 65\% and lowers inference latency, while preserving over 90\% of its original accuracy.
    \item \emph{LLaMA-3.2-11B (vision-Instruct)} \cite{llama3.1}: an open‐source vision‐capable LLM with an integrated image encoder. When quantized via Ollama, it achieves similar memory and speed improvements, enabling seamless single‐GPU deployment.
  \end{itemize}

\end{itemize}

\subsection{Baseline Framework}
To evaluate the performance and efficacy of our proposed method, we establish a robust baseline using a conventional RAG pipeline:
\begin{itemize}[nosep,leftmargin=10pt]
  \item \textbf{Basic RAG Setup:} Standard retrieval-augmented generation pipeline using fixed-size text chunks without semantic merging.
  \item \textbf{Embedding \& Retrieval:} Uses the same BAAI/bge-base-en-v1.5 embeddings as MultiFinRAG, but retrieves from fixed chunks via FAISS IVF-PQ without any tiered logic.
  \item \textbf{No Multimodal Parsing:} Charts and tables are not explicitly summarized: Visual and tabular elements are flattened into raw text or ignored, with no structured JSON conversion or image captioning.
\end{itemize}

\subsection{Proposed MultiFinRAG System}
\subsubsection{System Overview}

Figure~\ref{fig:overview}
illustrates our pipeline. Each PDF \(F_i\) is segmented into three sets of retrievable chunks:
\[
C_i \;=\; C_i^{\mathrm{text}} \;\cup\; C_i^{\mathrm{table}} \;\cup\; C_i^{\mathrm{image}},
\]
where \(C_i^{\mathrm{text}}\) are semantically coherent text passages, and \(C_i^{\mathrm{table}}\), \(C_i^{\mathrm{image}}\) are table and figure regions converted via a multimodal LLM. All chunks are embedded and stored in an approximate FAISS index. A query \(Q\) triggers a tiered retrieval (text‐only then text + table and image), automatically escalating whenever context is insufficient, before a final LLM answer generation.

\subsubsection{Semantic Chunking \& Indexing}

\paragraph{To avoid arbitrary splits and capture coherent semantic units, we:}

\begin{enumerate}[nosep,leftmargin=*,label=(\arabic*)]
  \item \textbf{Sentence segmentation:} split narrative into sentences \(S=\{s_1,\dots,s_n\}\).
  \item \textbf{Sliding windows:} with window size \(w\) and overlap \(o\), form overlapping blocks
    \[
      B_i = \{s_i,\dots,s_{i+w-1}\},\quad i=1,\,1+(w-o),\,\dots
    \]
  \item \textbf{Embedding \& breakpoints:} embed each sentence \(e_j=E(s_j)\), compute
    \[
      d_j = 1 - \frac{e_j\cdot e_{j+1}}{\|e_j\|\|e_{j+1}\|},\quad j=1,\dots,w-1,
    \]
    and mark any \(d_j\) above the 95th percentile of \(\{d_j\}\) as a split point.
  \item \textbf{Chunk formation:} split each block \(B_i\) at its breakpoints into semantic chunks, collecting all into \(C^{\mathrm{text}}\).
  \item \textbf{Chunk merging:} compute pairwise cosine similarities among the resulting chunks, then greedily merge any pairs whose similarity exceeds a high threshold (e.g.\ 0.85) to reduce redundancy.
  \item \textbf{Approximate indexing:} embed the final set of chunks and build a FAISS HNSW (or IVF-PQ) index, enabling sub‐second \(k\)-NN lookups at scale \(\lvert C\rvert>10^5\).
\end{enumerate}
\paragraph{Context‐size reduction:}  
By grouping semantically coherent sentences and then merging near‐duplicate chunks, we cut the total number of chunks—and hence the total token count—sent to the LLM by roughly 40–60\% on average. This directly translates into lower computational costs, reduced latency, and more efficient end‐to‐end QA without sacrificing retrieval quality.

\begin{figure}[t]
  \centering
  \includegraphics[width=\linewidth]{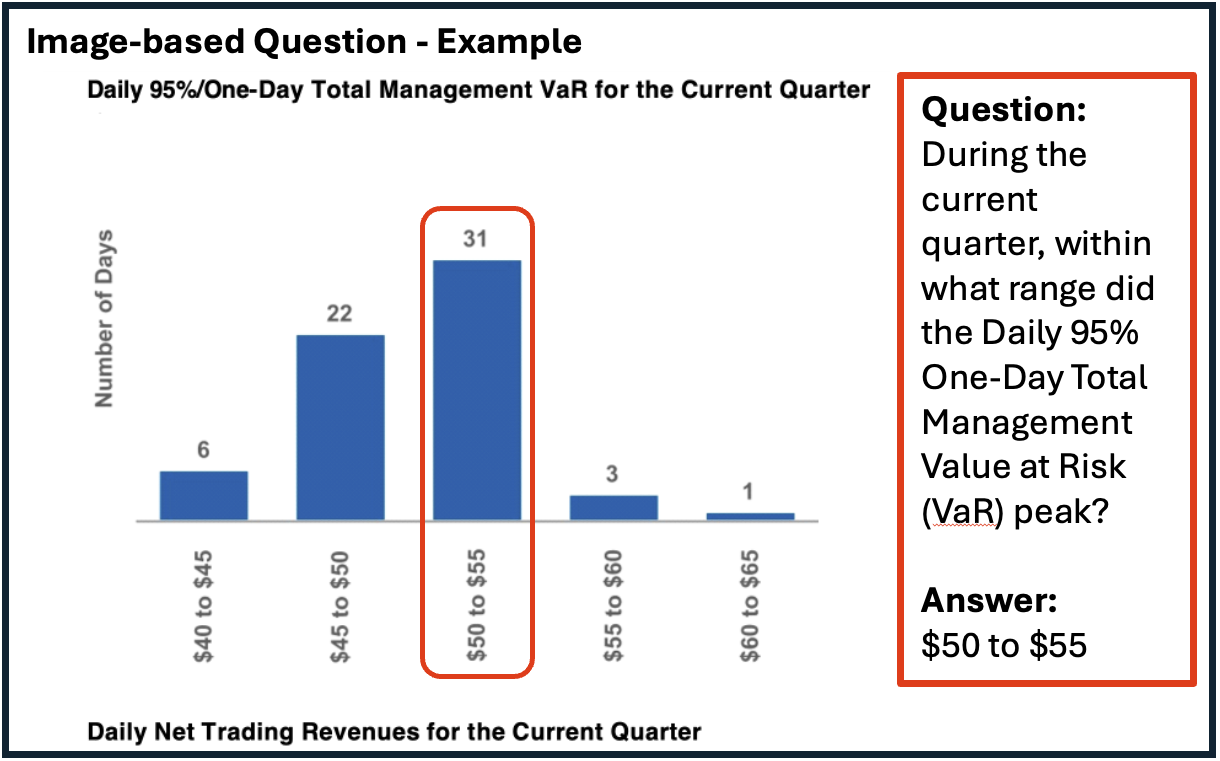}
  \caption{Example for type 2 questions}
  \label{type_2_ques}
  \Description{Example for type 2 questions}
\end{figure}

\subsubsection{Batch Multimodal Extraction}

Algorithm~\ref{alg:kb-build-sep} details our unified pipeline for ingesting tables and figures from each PDF.  After initial region detection (via Detectron2Layout and pdfminer) and semantic text chunking (lines 1–10), we process tables and images in two analogous, batched passes to maximize LLM throughput and ensure 100\% coverage:

\begin{algorithm}[h]
  \caption{Knowledge Base Construction }
  \label{alg:kb-build-sep}
  \begin{algorithmic}[1]
    \Require 
      PDF docs $\mathcal{F}$, batch size $B$, embedder $E$, 
      empty indexes $\mathcal{I}_{\text{text}},\mathcal{I}_{\text{table}},\mathcal{I}_{\text{image}}$
    \Ensure 
      Populated indexes per modality

    \ForAll{$F\in\mathcal{F}$}
      \State {\bfseries Region detection:}
      \State \quad Render pages $\to P_T,P_V$
      \State \quad Detect tables $T\gets\text{detect\_tables}(P_T)$
      \State \quad Detect figures $V\gets\text{detect\_figures}(P_V)$

      \State {\bfseries Text extraction \& semantic chunking:}
      \State \quad $X\gets\text{extract\_text}(F)$
      \State \quad $S\gets\text{segment\_sentences}(X)$
      \State \quad $C\gets\text{semantic\_chunk\_merge}(S)$
      \ForAll{$c\in C$}
        \State $e\gets E(c)$; normalize$(e)$
        \State $\mathcal{I}_{\text{text}}.\text{insert}(e,\{\,\text{content}:c\})$
      \EndFor

      \State {\bfseries Batch‐parse tables:}
      \State \quad Partition $T$ into $\lceil|T|/B\rceil$ batches
      \ForAll{batch $t_b$}
        \State $\{(desc_i,json_i)\}\gets\text{batch\_parse\_one\_batch}(t_b)$
        \ForAll{$(desc_i,json_i)$}
          \State $e\gets E(desc_i)$; normalize$(e)$
          \State $\mathcal{I}_{\text{table}}.\text{insert}\bigl(e,\{\,\text{summary}:desc_i,\text{json}:json_i\}\bigr)$
        \EndFor
      \EndFor

      \State {\bfseries Batch‐summarize images:}
      \State \quad Partition $V$ into $\lceil|V|/B\rceil$ batches
      \ForAll{batch $v_b$}
        \State $\{sum_j\}\gets\text{batch\_summarize}(v_b)$
        \ForAll{$sum_j$}
          \State $e\gets E(sum_j)$; normalize$(e)$
          \State $\mathcal{I}_{\text{image}}.\text{insert}(e,\{\text{summary}:sum_j\})$
        \EndFor
      \EndFor
    \EndFor
  \end{algorithmic}
\end{algorithm}

\vspace{0.5em}
\paragraph{Batch Table Parsing.}  
\begin{itemize}[nosep,leftmargin=1em]

\item Detected table regions \(T\) are first cropped with padding and saved as individual images.  We then partition \(T\) into \(\lceil|T|/B\rceil\) batches of size at most \(B\).  For each batch \(t_b\) (lines 11–17):
\item We construct a single multimodal prompt listing the exact filenames in \(t_b\) and invoke \texttt{batch\_parse\_one\_batch}, which returns \((\mathit{desc}_i,\mathit{json}_i)\) for each file.
\item Each textual description \(\mathit{desc}_i\) is embedded using the model \(E(\cdot)\) and inserted into the table index \(\mathcal{I}_{\mathrm{table}}\) along with its parsed JSON representation.
\item If any filename fails to appear in the response—due to an LLM omission or low confidence—we write a stub file in our dump directory and later retry with a single‐image prompt, guaranteeing no table goes unparsed.
\end{itemize}

\smallskip
\paragraph{Batch Image Summarization.}
\begin{itemize}[nosep,leftmargin=1em]
\item Similarly, figure regions \(V\) (charts, diagrams, etc.) are batched into \(\lceil|V|/B\rceil\) groups.  For each batch \(v_b\) (lines 18–23):
\item We send a batch prompt requesting a 3–6 sentence summary per image, explicitly instructing the LLM to ignore non–data visuals (e.g.\ logos, watermarks).
\item The returned summaries \(\{\mathit{sum}_j\}\) are embedded, normalized, and inserted into the image index \(\mathcal{I}_{\mathrm{image}}\).
\item Any missing summaries trigger a fallback to individual-image summarization, ensuring complete coverage of all data‐relevant figures.
\end{itemize}

By batching up to $B$ items per prompt, we amortize the LLM’s per‐call overhead and maintain exact filename\(\leftrightarrow\)output alignment, while the stub/fallback mechanism in each pass ensures no region is left unprocessed. All embeddings are stored in independent FAISS indexes for text, tables, and images, ready for downstream retrieval.

\subsubsection{Tiered Retrieval \& Decision Function}

Let \(n\), \(m\), and \(p\) denote the minimum number of text chunks required, the number of table chunks to fetch in the fallback, and the number of image summaries to fetch on fallback, respectively (we use \(n=6\), \(m=4\), \(p=3\)); these values were determined through trial and error on a holdout development set.

\begin{enumerate}[nosep, leftmargin=*, label=(\arabic*)]
  \item \textbf{Text‐only retrieval.}
    \[
      \mathcal{T}
      = \bigl\{\,c\in C^{\mathrm{text}}\;\bigm|\;
      \cosine\!\bigl(e_Q,\,E(c)\bigr)\ge\theta_{\mathrm{text}}
      \bigr\}.
    \]

    If \(\lvert\mathcal{T}\rvert \ge n\), issue a single LLM call with \(\mathcal{T}\) as context and terminate.

  \item \textbf{Table fallback.}
    \[
      \mathcal{T}_{\mathrm{tbl}}
      = \mathrm{Top}_{m}\bigl\{\,c\in C^{\mathrm{table}}\;\bigm|\;
      \cosine\!\bigl(e_Q,\,E(c)\bigr)\ge\theta_{\mathrm{table}}
      \bigr\}.
    \]

  \item \textbf{Image fallback.}
    \[
      \mathcal{T}_{\mathrm{img}}
      = \mathrm{Top}_{p}\bigl\{\,c\in C^{\mathrm{image}}\;\bigm|\;
      \cosine\!\bigl(e_Q,\,E(c)\bigr)\ge\theta_{\mathrm{image}}
      \bigr\}.
    \]

  \item \textbf{Combined prompt.}  
    Concatenate any non-empty sets among \(\mathcal{T}\), \(\mathcal{T}_{\mathrm{tbl}}\), and \(\mathcal{T}_{\mathrm{img}}\) (including each table’s JSON + summary), then issue the final LLM call.
\end{enumerate}

All retrievals use an approximate FAISS index, and each LLM call carries a system prompt instructing it to explicitly defer (“insufficient information”) rather than hallucinate. We instrument every stage with wall-clock timers and \texttt{tqdm} progress bars to monitor end-to-end performance.

\begin{figure}
  \centering
  \includegraphics[width=\linewidth]{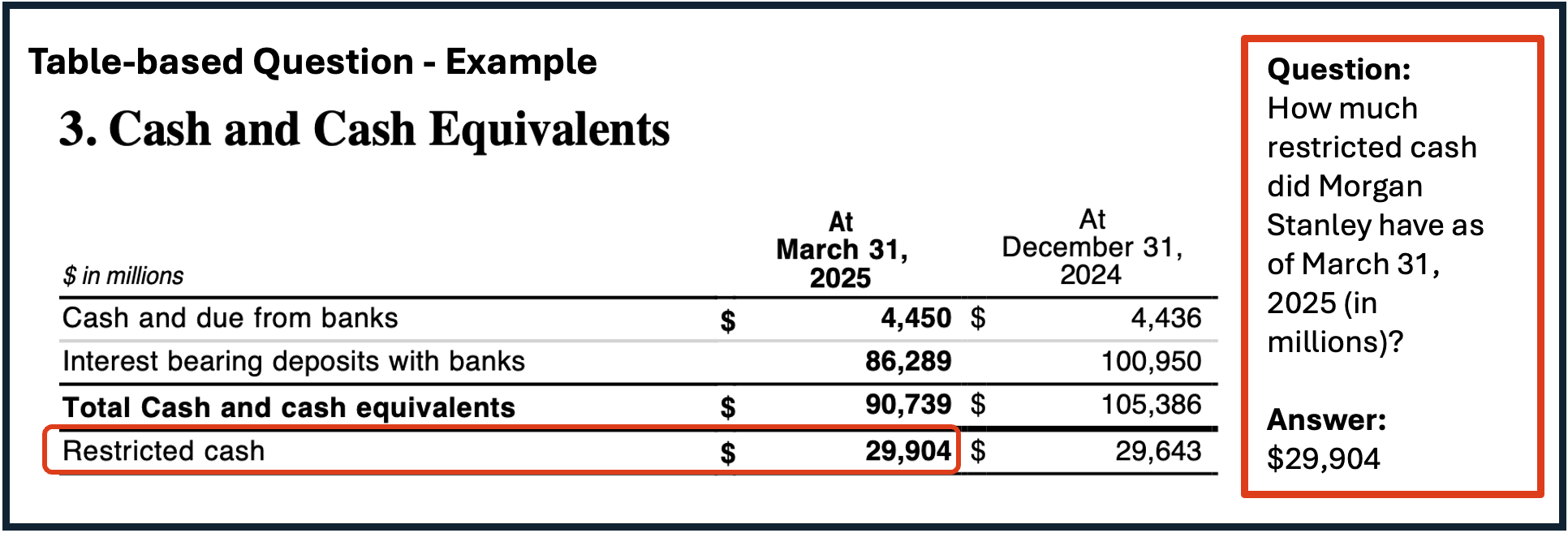}
  \caption{Example for type 3 questions}
  \label{type_3_ques}
  \Description{Example for type 3 questions}
\end{figure}

\begin{figure*}[t]
  \centering
  \includegraphics[width=\linewidth]{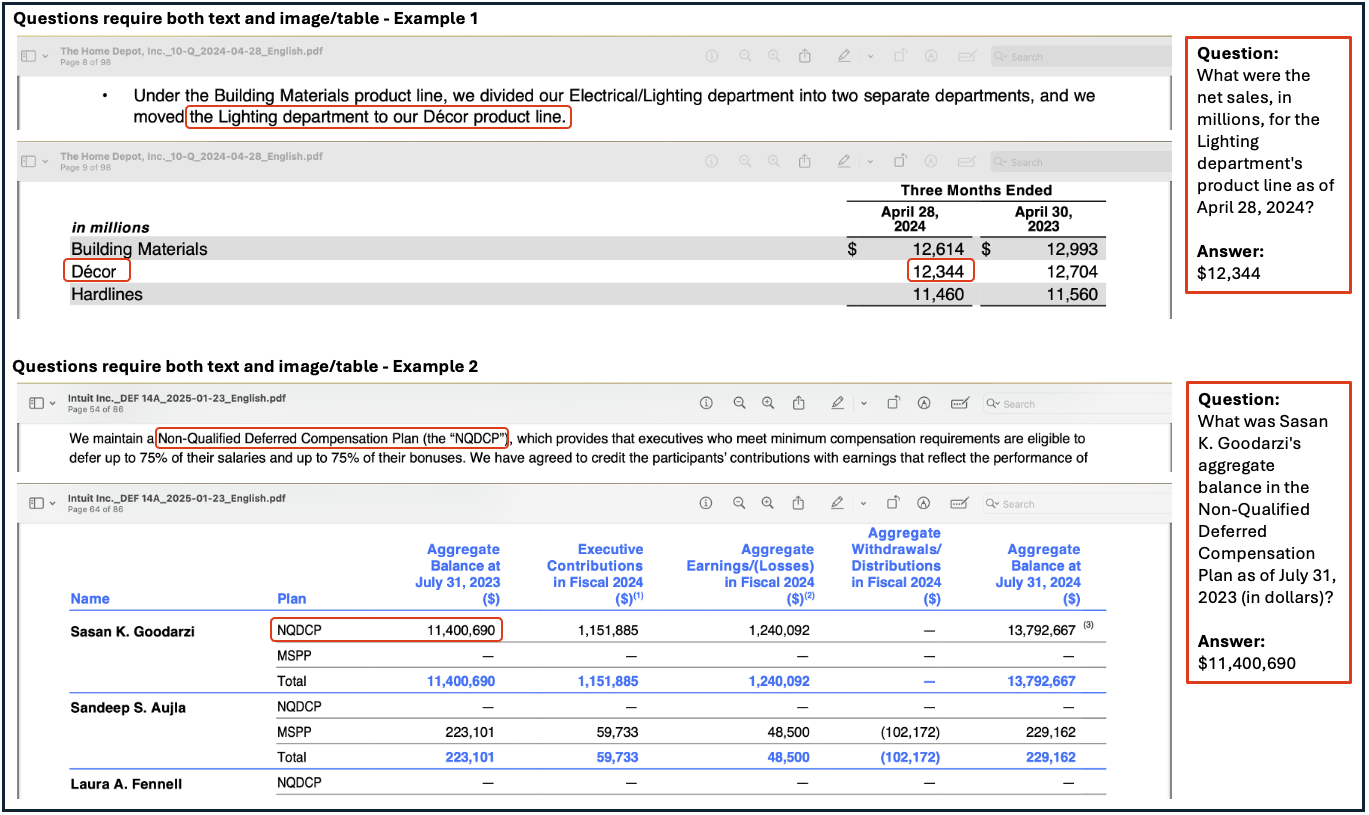}
  \caption{Examples for type 4 questions}
  \label{type_4_ques}
  \Description{Examples for type 4 questions}
\end{figure*}

\subsubsection{Threshold Calibration via Decision Function}
To select optimal similarity cut‐offs for each modality, we ran a decision function over a held‐out set of sample queries and measured both retrieval quality (precision of retrieved contexts) and end-to-end QA accuracy:
\vspace{2pt}
\begin{enumerate}[nosep,leftmargin=*,label=(\arabic*)]
  \item \textbf{Text threshold sweep:} Vary $\theta_{\text{text}}$ from 0.55 to 0.85 in steps of 0.05; at each value, retrieve all text chunks with $\cos(E(Q),E(c)) \ge \theta_{\text{text}}$, feed the top-$k$ to the LLM, and record answer accuracy.
  \item \textbf{Table \& image threshold sweep:} Keeping $\theta_{\text{text}}$ fixed, vary $(\theta_{\text{table}},\theta_{\text{image}})$ independently from 0.55 to 0.75; at each pair, retrieve table and image summaries above their thresholds, query the LLM, and record accuracy.
  \item \textbf{Decision criterion:} Choose the triplet $(\theta_{\text{text}},\theta_{\text{table}},\theta_{\text{image}})$ that maximized a combined metric of context relevance and QA accuracy, while keeping per-query context size under budget.
\end{enumerate}

\vspace{2pt}
The final thresholds selected by this procedure were
\[
  \theta_{\text{text}} = 0.70,\quad
  \theta_{\text{table}} = 0.65,\quad
  \theta_{\text{image}} = 0.55.
\]
These values balance precision and coverage, ensuring text‐only retrieval succeeds when possible, and gracefully falling back to richer table/image contexts when needed.

\section{Evaluation}

\paragraph{Note on Retrieval Metrics.}
This study focuses on end‐to‐end QA accuracy. Detailed information‐retrieval statistics (precision, recall, etc.) for the retrieval components were collected during development but are omitted here due to space constraints; we plan to report them in a follow‐up paper.

\begin{table*}[t]
  \caption{MultiFinRAG Performance Comparison Across Question Types}
  \label{tab:evaluation_table}
  \begin{tabular}{lrrrrrr}
    \toprule
    Question & Number of & Baseline   & Baseline   & MultiFinRAG & \textbf{MultiFinRAG} & ChatGPT-4o\\
    Type     & Questions & with Llama & with Gemma & with Llama  & \textbf{with Gemma}          & (Free Tier)\\
    \midrule
    1. Text-based             & 146  & 75.3\% & 76.7\% & 83.6\% & \textbf{90.4\%} & 86.3\% \\
    2. Image-based            &  42  &    0\% &    0\% & 42.9\% & \textbf{66.7\%} & 23.8\% \\
    3. Table-based            &  72  &  2.8\% &  5.6\% & 13.9\% & \textbf{69.4\%} & 44.4\% \\
    4. Text+Image/Table-based &  40  &    0\% &    0\% & 10.0\% & \textbf{40.0\%} & 15.0\% \\
    Total                     & 300  & 33.3\% & 36.7\% & 47.3\% & \textbf{75.3\%} & 56.0\% \\
    \bottomrule
  \end{tabular}
\end{table*}

\subsection{Dataset}
For our evaluation, we collected financial documents from various companies, including Form 10-Q (Quarterly Report), Form 10-K (Annual Report), Form 8-K with EX-99.1 (Current Report), and DEF 14A (Proxy Statement). These documents were obtained by accessing the SEC’s EDGAR database at \url{https://www.sec.gov/edgar/search/} and utilizing the SEC API to download and convert the forms into PDF files. We then manually crafted questions to ensure they met our expected difficulty levels and quality standards.

To ensure that the pipeline accurately comprehends the files, we developed four distinct types of questions with varying levels of difficulty. The simplest are text-based questions, followed by image-based and table-based questions. The most challenging category involves questions that require the interpretation of both text and images or tables. Details of each question type are provided below.

Question Distribution - We created a total of 300 evaluation questions, distributed as follows (see Table~\ref{tab:evaluation_table}):
\begin{itemize}[nosep,leftmargin=*]
  \item 146 text-based
  \item 42 image-based
  \item 72 table-based
  \item 40 requiring both text and image/table reasoning
\end{itemize}

\subsubsection{Text-based Questions}
This type of questions are straightforward and can be answered using the text in the financial documents. For example, we use the sentence "A weaker U.S. dollar positively impacted net sales by \$106 million during the first quarter of fiscal 2024." from the Form 10-Q of The Home Depot, Inc. to create the question: "How much did the weaker U.S. dollar positively impact net sales during the first quarter of fiscal 2024 (in millions)?" The answer is "106." We ensure that the answer is a value, term, or a few words to reduce evaluation confusion.

\subsubsection{Image-based questions}
This type of questions require analyzing images and graphs in financial documents. These questions test the framework's ability to understand financial graphs. For example, pie charts use different colors to represent various categories. We will create a question about categories to assess whether the MultiFinRAG framework can capture such details. Figure \ref{type_2_ques} illustrates another example: using the bar chart from Morgan Stanley's Form 10-Q, we create the question, "During the current quarter, within what range did the Daily 95\% One-Day Total Management Value at Risk (VaR) peak?" The correct answer is "\$50 to \$55." The framework needs to capture the value for each bar and identify the bar with the highest value to provide the correct answer. When crafting the questions and answers, we ensure that the answers do not appear elsewhere in the document.

\subsubsection{Table-based questions}
This type of questions involve interpreting tables in financial documents. These questions test the framework's ability to locate and understand tables. For instance, in Figure \ref{type_3_ques}, using a table from Morgan Stanley's Form 10-Q, we formulate the question: "How much restricted cash did Morgan Stanley have as of March 31, 2025 (in millions)?" The answer is "29,904." To correctly answer this question, the framework needs to identify the appropriate row and column to retrieve the value. Similar to image-based questions, we search for the value "29,904" in the file and ensure it is not mentioned in the textual content.

\subsubsection{Questions require both text and image/table}
Questions require both text and image/table are the most challenging, as they require information from both text and images or tables. For example, in Figure \ref{type_4_ques}, the first example is: "What were the net sales, in millions, for the Lighting department’s product line as of April 28, 2024?" To answer this, one must first identify that the Lighting department’s product line is "Decor" from the text, and then refer to the related table to find that the value for the Decor product line is "12,344." 

Another example from Figure \ref{type_4_ques} is: "What was Sasan K. Goodarzi’s aggregate balance in the Non-Qualified Deferred Compensation Plan as of July 31, 2023 (in dollars)?" This requires recognizing that the Non-Qualified Deferred Compensation Plan is abbreviated as "NQDCP" in the table, and then locating that Sasan K. Goodarzi’s NQDCP balance at the specified time was "\$11,400,690."

\subsection{Evaluation Strategy}
Evaluating the output of LLM frameworks remains a significant open research area. Traditional methods that assess whether the generated answer exactly matches the reference answer are increasingly unsuitable for evaluating LLM results. This is because it is challenging to ensure that generated answers precisely replicate reference answers. For instance, even when most answers in our dataset are simple numeric values, discrepancies in units can arise, such as a reference answer of "1 billion" versus a generated answer of "1,000 million."

Another popular evaluation method, BERTScore \cite{BERTScore}, computes a similarity score between the generated answer and the reference answer. However, this approach is not suitable for our study. This limitation arises because many of the questions we create pertain to numerical values, making it illogical to embed numbers and compare their semantic similarities.

Recently, there has been an increasing interest in evaluating GenAI results using LLMs \cite{LLM-Assisted_VQA_Evaluation, G-eval}. However, these studies all highlight a downside: the inherent bias of the LLMs used for the evaluation tasks. While LLMs can be employed for evaluating large volumes of QA pairs, the size of our dataset and resource constraints lead us to opt for manual evaluations. This approach ensures more accurate assessments and minimizes the risk of misjudgments.


\subsection{Performance Comparison: Baselines vs. MultiFinRAG}

\subsubsection{Text-based Questions} When examining the accuracy for text-based questions in Table \ref{tab:evaluation_table}, two key observations emerge:

Firstly, there is a significant improvement in accuracy when transitioning from the Baseline framework to MultiFinRAG. For example, the accuracy for text-based questions using MultiFinRAG with Gemma 3 is 90.4\%, representing an improvement of 15.1\% compared to the Baseline framework with Gemma 3 (75.3\%). This increase indicates that embedding refined chunks into an IVF/PQ FAISS index, which enables sub-second approximate k-NN lookup at scale, is more effective than the baseline approach that retrieves fragmented or redundant passages.

Secondly, Gemma 3 outperforms Llama-3.2-11B-Vision-Instruct across both Baseline and MultiFinRAG frameworks. Specifically, MultiFinRAG with Gemma 3 achieves 90.4\% accuracy for text-based questions, which is 6.8\% higher than MultiFinRAG with Llama-3.2-11B-Vision-Instruct (83.6\%). This suggests that Gemma 3 is more effective at identifying key answers from the retrieval trunks once the LLMs receive them.

\subsubsection{Image-based Questions} The accuracy difference for image-based questions between the Baseline framework and MultiFinRAG demonstrates that the Baseline framework cannot effectively extract details from images, as anticipated. This limitation arises because the Baseline framework only truncates and embeds text for retrieving answers, resulting in weaker performance for both Baseline with Llama-3.2-11B-Vision-Instruct and Baseline with Gemma 3 in this question type.

In contrast, MultiFinRAG with Gemma 3 achieves higher accuracy (66.7\%) compared to MultiFinRAG with Llama-3.2-11B-Vision-Instruct (42.9\%), indicating that Gemma 3 is more adept at describing images and graphs in the files in a detailed and comprehensive manner.

\subsubsection{Table-based Questions} Similar to image-based questions, the Baseline framework performs poorly on table-based questions. Although the Baseline framework may occasionally answer correctly for questions based on relatively simple table structures, it generally fails to provide accurate responses for this question type.

Moreover, when applying different LLMs to describe tables, Gemma 3 provides significantly better descriptions than Llama-3.2-11B-Vision-Instruct. This results in a 55.5\% higher accuracy for MultiFinRAG with Gemma 3 (69.4\%) compared to MultiFinRAG with Llama-3.2-11B-Vision-Instruct (13.9\%).

\subsubsection{Questions Requiring Both Text and Image/Table} Table \ref{tab:evaluation_table} illustrates that the accuracy for questions requiring both text and image/table analysis varies significantly between MultiFinRAG with Llama-3.2-11B-Vision-Instruct and MultiFinRAG with Gemma 3. This variance highlights the importance of the LLMs’ ability to accurately describe graphs and tables within the MultiFinRAG framework. Specifically, using MultiFinRAG with is Llama-3.2-11B-Vision-Instruct too long to use? yields only 10\% accuracy for these complex questions, whereas using MultiFinRAG with Gemma 3 achieves 40\% accuracy.

Additionally, the transition from the Baseline to the MultiFinRAG framework notably improves the accuracy rate. For instance, accuracy increases from 0\% with the Baseline framework using Gemma 3 to 40\% with MultiFinRAG using Gemma 3. This improvement underscores the MultiFinRAG framework's ability to retrieve relevant trunks and integrate information effectively to generate accurate final answers.

\begin{figure}[t]
  \centering
  \fbox{%
  \includegraphics[width=\linewidth]{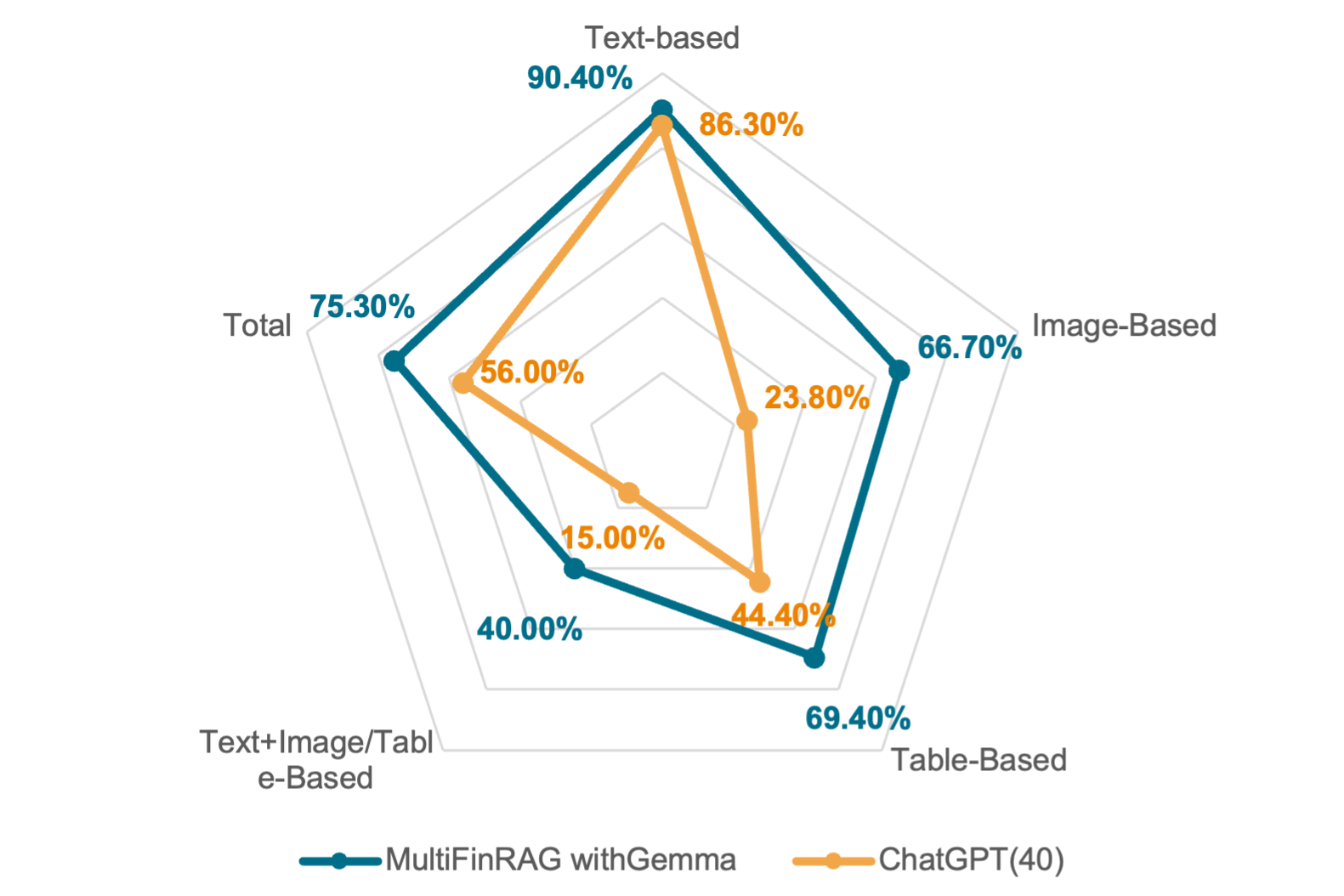}
  }
  \caption{Accuracy Comparison by Question Type}
  \label{Benchmarking_against_chatgpt}
  \Description{Accuracy Comparison by Question Type}
\end{figure}


\subsection{Benchmarking against ChatGPT-4o}

We compare the performance of MultiFinRAG against ChatGPT-4o across multiple question types. ChatGPT-4o \cite{gpt4o} serves as a strong benchmark due to its advanced general-purpose reasoning and wide availability in the free tier, making it a relevant standard for evaluating our domain-specific, multimodal QA system.

\subsubsection{Text-based Questions}
In Figure \ref{Benchmarking_against_chatgpt}, we observe that compared to ChatGPT-4o, MultiFinRAG with Gemma 3 demonstrates similar accuracy, being only 4.1\% higher. However, we identified instances where MultiFinRAG with Gemma 3 provided correct answers while ChatGPT-4o did not. For example, consider the financial statement:

"As of March 31, 2025, and December 31, 2024, 98\% of the Firm’s portfolio of HTM securities were investment-grade U.S. agency securities, U.S. Treasury securities, and Agency CMBS, which were on accrual status and assumed to have zero credit losses."

The correct answer is "98\%." In this case, ChatGPT-4o incorrectly responded with "100\%," likely due to the presence of multiple relevant values within the document.

\subsubsection{Image-based Questions}
MultiFinRAG with Gemma 3 achieves over 40\% higher accuracy than ChatGPT-4o for image-based questions. For instance, in an image-based question depicted in Figure \ref{type_2_ques}, the correct answer is "\$50 to \$55," whereas ChatGPT-4o inaccurately responded with "\$63 to \$98." This confusion likely arises from the presence of multiple values (\$63 and \$98) in the text and tables within the file, leading to incorrect information retrieval by ChatGPT-4o.

\subsubsection{Table-based Questions}
When addressing table-based questions, MultiFinRAG with Gemma 3 outperforms ChatGPT-4o by over 25\% in accuracy. For example, in Figure \ref{type_3_ques}, the ChatGPT-4o incorrectly stated:

"As of March 31, 2025, Morgan Stanley had \$0 million in restricted cash. This is explicitly stated in the report: no amount is separately listed or disclosed under restricted cash, indicating that it was not material or not applicable at that reporting date."

However, this information is not expressed anywhere in the file, suggesting that ChatGPT-4o's response may be influenced by its existing training data rather than the provided document.

\subsubsection{Questions Requiring Both Text and Image/Table}
Finally, for questions requiring both text and image/table analysis, MultiFinRAG with Gemma 3 achieved an accuracy of 75.3\%, compared to ChatGPT-4o's 56.0\%, marking a 19.3\% improvement. We observed that when the necessary information is already present in ChatGPT-4o's existing knowledge base, it can answer correctly. For example, in Figure \ref{type_4_ques}, Example 2, ChatGPT-4o correctly identified that the "Non-Qualified Deferred Compensation Plan" is often abbreviated as "NQDCP" and applied this knowledge to answer the question accurately.

However, in cases where the required information is not familiar or does not exist in its knowledge base, such as Example 1 in Figure \ref{type_4_ques}, ChatGPT-4o struggled to integrate the information effectively, resulting in incorrect or incomplete answers.

\subsection{Efficiency and Cost}
In this evaluation, we primarily assess accuracy rather than processing time, as the latter is significantly influenced by the underlying infrastructure's capabilities, such as computational resources and system configurations. The average processing time for MultiFinRAG with Gemma 3 is approximately 25 minutes when handling a financial file consisting of a 200-page PDF with 200 tables and 150 images on Google Colab \cite{GoogleColab} T4 GPU which has 16 GB RAM \cite{d2lGPUarch}.

Regarding cost, both the baseline and MultiFinRAG systems can be operated using Google Colab's free tier at the time of writing. Our experiments were conducted using this free version, resulting in no associated expenses. Similarly, the results obtained from ChatGPT-4o, used as a comparative benchmark, were achieved using its free tier, incurring no additional costs.

\section{Discussion and Future Work}
MultiFinRAG demonstrates strong performance and efficiency in extracting insights from complex financial documents. However, several avenues remain open for future enhancements, as mentioned below:
\begin{itemize}[nosep,leftmargin=*]
  \item \textbf{Module-wise evaluation:} Systematic ablations (e.g., disabling batch extraction or tiered fallback) will help quantify each component’s individual impact.
 
  \item \textbf{User feedback scope:} All 300 QA pairs were manually verified by our team of domain experts, ensuring answer correctness; future work will include broader user studies to assess usability and workflow integration.

  \item \textbf{Structured‐Data Pipeline:}  
    To handle large tabular attachments (e.g.\ CSV/Excel) beyond our current JSON‐and‐summary stage, we are prototyping an ingestion‐and‐query system. It will normalize heterogeneous table formats into a lightweight database, expose a natural‐language interface for precise data retrieval, and feed the resulting rows and aggregates directly into our RAG generator as structured context.

  \item \textbf{Cross‐Document \& Longitudinal Analysis:}  
    Many financial queries require comparisons over multiple time periods or across related filings (e.g., 10‐K vs.\ 10‐Q). Future work will:
    \begin{itemize}[nosep,leftmargin=*]
      \item Build a unified multi‐document index supporting temporal joins and trend detection;
      \item Develop specialized retrieval that pulls “paired” chunks (e.g., Q1 vs.\ Q2) and surfaces year‐over‐year deltas;
      \item Introduce timeline visualizations and automated narrative summaries of longitudinal changes.
    \end{itemize}

  \item \textbf{Robustness to Noise \& Error Correction:}  
    PDF parsing and OCR remain brittle in the face of low‐quality scans or unusual layouts. We intend to:
    \begin{itemize}[nosep,leftmargin=*]
      \item Ensemble multiple OCR/layout engines to triangulate cell boundaries and text;
      \item Apply consistency checks (e.g., row‐sum invariants, unit‐sanity) to detect mis‐parsed numbers;
      \item Use small fine‐tuned correction LLMs to post‐process and validate extracted tables.
    \end{itemize}

  \item \textbf{Extended Domain Coverage:}  
    Beyond annual reports, our methods can generalize to other financial and regulatory documents—S‐1 filings, prospectuses, and earnings‐call transcripts. Each domain brings new layout conventions, metadata fields, and entity types that we plan to accommodate via modular extraction components.

  \item \textbf{Fine‐Tuning \& Domain Adaptation:}  
    Finally, we plan to fine‐tune both our retrieval and generation models on proprietary, high‐quality financial QA datasets, exploring supervised contrastive training for retrieval and very‐low‐rank adapters (LoRA) for generator customization.

  \item \textbf{Web‐Article Ingestion \& Real‐Time Multimodal Q\&A:}  
    While MultiFinRAG is optimized for long‐form PDF filings, many financial insights also reside in online news articles and blogs, which often include embedded tables and charts. We propose extending our pipeline to live web content by:
    \begin{itemize}[nosep,leftmargin=*]
    \item Render and segment HTML via a headless browser (e.g.\ Puppeteer).
    \item Extract tables using DOM parsing (e.g.\ \texttt{pandas.read\_html()}) with screenshot + OCR fallback.
    \item Summarize figures (\texttt{<img>}, \texttt{<svg>}) in batches.
    \item Integrate all modalities—text chunks, table JSON, figure summaries—into the FAISS-based retrieval and generation pipeline, optionally pre-filtered by a FANAL classifier \cite{patel2024fanalfinancialactivity} or CANAL-style filter \cite{Patel_2024}.
  \end{itemize}
    This extension would transform MultiFinRAG into a web‐scale, real‐time multimodal Q\&A engine—bringing its precision retrieval and cross‐modal reasoning to the dynamic realm of online financial news.
\end{itemize}

These future directions aim to evolve MultiFinRAG from a PDF‐centric RAG pipeline into an interactive, fully‐featured financial intelligence platform—capable of handling massive tables, cross‐document trends, real‐time updates, and robust error correction, all under user‐friendly natural‐language control.

\section{Conclusion}

We demonstrated that MultiFinRAG delivers precise answers to complex financial queries from extensive multimodal PDF filings, surpassing ChatGPT-4o in accuracy on questions involving text, tables, and images. By combining modality-aware retrieval thresholds with lightweight, quantized open-source LLMs, the framework operates efficiently on modest hardware—reducing token usage by over 60\% and accelerating response times. With over 75\% accuracy on challenging multimodal QA tasks, MultiFinRAG provides a practical, scalable, and cost-effective solution for querying large, information-rich financial documents through advanced multimodal reasoning.

\appendix
\newpage
\bibliographystyle{ACM-Reference-Format}
\bibliography{references}


\begin{thebibliography}{31}


\ifx \showCODEN    \undefined \def \showCODEN     #1{\unskip}     \fi
\ifx \showDOI      \undefined \def \showDOI       #1{#1}\fi
\ifx \showISBNx    \undefined \def \showISBNx     #1{\unskip}     \fi
\ifx \showISBNxiii \undefined \def \showISBNxiii  #1{\unskip}     \fi
\ifx \showISSN     \undefined \def \showISSN      #1{\unskip}     \fi
\ifx \showLCCN     \undefined \def \showLCCN      #1{\unskip}     \fi
\ifx \shownote     \undefined \def \shownote      #1{#1}          \fi
\ifx \showarticletitle \undefined \def \showarticletitle #1{#1}   \fi
\ifx \showURL      \undefined \def \showURL       {\relax}        \fi
\providecommand\bibfield[2]{#2}
\providecommand\bibinfo[2]{#2}
\providecommand\natexlab[1]{#1}
\providecommand\showeprint[2][]{arXiv:#2}

\bibitem[AI(2024)]%
        {llama3.1}
\bibfield{author}{\bibinfo{person}{Meta AI}.} \bibinfo{year}{2024}\natexlab{}.
\newblock \bibinfo{title}{Introducing Llama 3.1: Our most capable models to date}.
\newblock \bibinfo{howpublished}{\url{https://ai.meta.com/blog/meta-llama-3-1/}}.
\newblock
\newblock
\shownote{Accessed: December 18, 2024}.


\bibitem[Asai et~al\mbox{.}(2023)]%
        {selfrag}
\bibfield{author}{\bibinfo{person}{Akari Asai}, \bibinfo{person}{Zeqiu Wu}, \bibinfo{person}{Yizhong Wang}, \bibinfo{person}{Avirup Sil}, {and} \bibinfo{person}{Hannaneh Hajishirzi}.} \bibinfo{year}{2023}\natexlab{}.
\newblock \bibinfo{title}{Self-RAG: Learning to Retrieve, Generate, and Critique through Self-Reflection}.
\newblock
\newblock
\showeprint[arxiv]{2310.11511}~[cs.CL]
\urldef\tempurl%
\url{https://arxiv.org/abs/2310.11511}
\showURL{%
\tempurl}


\bibitem[Dong et~al\mbox{.}(2024)]%
        {dpa}
\bibfield{author}{\bibinfo{person}{Guanting Dong}, \bibinfo{person}{Yutao Zhu}, \bibinfo{person}{Chenghao Zhang}, \bibinfo{person}{Zechen Wang}, \bibinfo{person}{Zhicheng Dou}, {and} \bibinfo{person}{Ji-Rong Wen}.} \bibinfo{year}{2024}\natexlab{}.
\newblock \bibinfo{title}{Understand What LLM Needs: Dual Preference Alignment for Retrieval-Augmented Generation}.
\newblock
\newblock
\showeprint[arxiv]{2406.18676}~[cs.CL]
\urldef\tempurl%
\url{https://arxiv.org/abs/2406.18676}
\showURL{%
\tempurl}


\bibitem[Douze et~al\mbox{.}(2024)]%
        {douze2024faiss}
\bibfield{author}{\bibinfo{person}{Matthijs Douze}, \bibinfo{person}{Alexandr Guzhva}, \bibinfo{person}{Chengqi Deng}, \bibinfo{person}{Jeff Johnson}, \bibinfo{person}{Gergely Szilvasy}, \bibinfo{person}{Pierre-Emmanuel Mazaré}, \bibinfo{person}{Maria Lomeli}, \bibinfo{person}{Lucas Hosseini}, {and} \bibinfo{person}{Hervé Jégou}.} \bibinfo{year}{2024}\natexlab{}.
\newblock \showarticletitle{The Faiss library}.
\newblock \bibinfo{journal}{\emph{arXiv preprint}}  \bibinfo{volume}{1} (\bibinfo{year}{2024}), \bibinfo{numpages}{xx}~pages.
\newblock
\showeprint[arxiv]{2401.08281}~[cs.LG]


\bibitem[Fatehkia et~al\mbox{.}(2024)]%
        {trag}
\bibfield{author}{\bibinfo{person}{Masoomali Fatehkia}, \bibinfo{person}{Ji~Kim Lucas}, {and} \bibinfo{person}{Sanjay Chawla}.} \bibinfo{year}{2024}\natexlab{}.
\newblock \bibinfo{title}{T-RAG: Lessons from the LLM Trenches}.
\newblock
\newblock
\showeprint[arxiv]{2402.07483}~[cs.AI]
\urldef\tempurl%
\url{https://arxiv.org/abs/2402.07483}
\showURL{%
\tempurl}


\bibitem[Google(2023)]%
        {GoogleColab}
\bibfield{author}{\bibinfo{person}{Google}.} \bibinfo{year}{2023}\natexlab{}.
\newblock \bibinfo{title}{Google Colaboratory}.
\newblock \bibinfo{howpublished}{\url{https://colab.research.google.com/}}.
\newblock
\newblock
\shownote{Accessed: May 15, 2025}.


\bibitem[Günther et~al\mbox{.}(2024)]%
        {gunther2024latechunking}
\bibfield{author}{\bibinfo{person}{Michael Günther} {et~al\mbox{.}}} \bibinfo{year}{2024}\natexlab{}.
\newblock \showarticletitle{Late Chunking: Contextual Chunk Embeddings Using Long-Context Embedding Models}.
\newblock \bibinfo{journal}{\emph{arXiv preprint arXiv:2401.12345}} \bibinfo{volume}{1}, \bibinfo{number}{1} (\bibinfo{year}{2024}), \bibinfo{pages}{xx--yy}.
\newblock


\bibitem[Kannammal et~al\mbox{.}(2025)]%
        {kannammal2025fin}
\bibfield{author}{\bibinfo{person}{KE Kannammal}, \bibinfo{person}{Mr~Anirudh RK}, \bibinfo{person}{Kuzhali Tamizhiniyal~P}, {et~al\mbox{.}}} \bibinfo{year}{2025}\natexlab{}.
\newblock \showarticletitle{Fin-Rag A Rag System for Financial Documents}.
\newblock \bibinfo{journal}{\emph{International Journal of Innovative Science and Research Technology}} \bibinfo{volume}{10}, \bibinfo{number}{4} (\bibinfo{year}{2025}), \bibinfo{pages}{1761--1767}.
\newblock


\bibitem[Karpukhin et~al\mbox{.}(2020)]%
        {dsp}
\bibfield{author}{\bibinfo{person}{Vladimir Karpukhin}, \bibinfo{person}{Barlas Oğuz}, \bibinfo{person}{Sewon Min}, \bibinfo{person}{Patrick Lewis}, \bibinfo{person}{Ledell Wu}, \bibinfo{person}{Sergey Edunov}, \bibinfo{person}{Danqi Chen}, {and} \bibinfo{person}{Wen tau Yih}.} \bibinfo{year}{2020}\natexlab{}.
\newblock \bibinfo{title}{Dense Passage Retrieval for Open-Domain Question Answering}.
\newblock
\newblock
\showeprint[arxiv]{2004.04906}~[cs.CL]
\urldef\tempurl%
\url{https://arxiv.org/abs/2004.04906}
\showURL{%
\tempurl}


\bibitem[Lewis et~al\mbox{.}(2020)]%
        {lewis2020rag}
\bibfield{author}{\bibinfo{person}{Patrick Lewis}, \bibinfo{person}{Ethan Perez}, \bibinfo{person}{Aleksandra Piktus}, \bibinfo{person}{Fabio Petroni}, \bibinfo{person}{Vladimir Karpukhin}, \bibinfo{person}{Naman Goyal}, \bibinfo{person}{Heinrich K\"{u}ttler}, \bibinfo{person}{Mike Lewis}, \bibinfo{person}{Wen tau Yih}, \bibinfo{person}{Tim Rockt\"{a}schel}, \bibinfo{person}{Sebastian Riedel}, {and} \bibinfo{person}{Douwe Kiela}.} \bibinfo{year}{2020}\natexlab{}.
\newblock \showarticletitle{Retrieval-Augmented Generation for Knowledge-Intensive NLP Tasks}. In \bibinfo{booktitle}{\emph{Advances in Neural Information Processing Systems (NeurIPS)}}. \bibinfo{publisher}{Curran Associates, Inc.}
\newblock
\urldef\tempurl%
\url{https://proceedings.neurips.cc/paper/2020/file/6b493230205f780e1bc26945df7481e5-Paper.pdf}
\showURL{%
\tempurl}


\bibitem[Li et~al\mbox{.}(2020)]%
        {li2020tablebankbenchmarkdatasettable}
\bibfield{author}{\bibinfo{person}{Minghao Li}, \bibinfo{person}{Lei Cui}, \bibinfo{person}{Shaohan Huang}, \bibinfo{person}{Furu Wei}, \bibinfo{person}{Ming Zhou}, {and} \bibinfo{person}{Zhoujun Li}.} \bibinfo{year}{2020}\natexlab{}.
\newblock \bibinfo{title}{TableBank: A Benchmark Dataset for Table Detection and Recognition}.
\newblock
\newblock
\showeprint[arxiv]{1903.01949}~[cs.CV]
\urldef\tempurl%
\url{https://arxiv.org/abs/1903.01949}
\showURL{%
\tempurl}


\bibitem[Liu et~al\mbox{.}(2024)]%
        {liu2024deepseek}
\bibfield{author}{\bibinfo{person}{Aixin Liu}, \bibinfo{person}{Bei Feng}, \bibinfo{person}{Bing Xue}, \bibinfo{person}{Bingxuan Wang}, \bibinfo{person}{Bochao Wu}, \bibinfo{person}{Chengda Lu}, \bibinfo{person}{Chenggang Zhao}, \bibinfo{person}{Chengqi Deng}, \bibinfo{person}{Chenyu Zhang}, \bibinfo{person}{Chong Ruan}, {et~al\mbox{.}}} \bibinfo{year}{2024}\natexlab{}.
\newblock \showarticletitle{Deepseek-v3 technical report}.
\newblock \bibinfo{journal}{\emph{arXiv preprint arXiv:2412.19437}} (\bibinfo{year}{2024}), \bibinfo{pages}{xx--yy}.
\newblock


\bibitem[Liu et~al\mbox{.}(2023)]%
        {G-eval}
\bibfield{author}{\bibinfo{person}{Yang Liu}, \bibinfo{person}{Dan Iter}, \bibinfo{person}{Yichong Xu}, \bibinfo{person}{Shuohang Wang}, \bibinfo{person}{Ruochen Xu}, {and} \bibinfo{person}{Chenguang Zhu}.} \bibinfo{year}{2023}\natexlab{}.
\newblock \bibinfo{title}{G-Eval: NLG Evaluation using GPT-4 with Better Human Alignment}.
\newblock
\newblock
\showeprint[arxiv]{2303.16634}~[cs.CL]
\urldef\tempurl%
\url{https://arxiv.org/abs/2303.16634}
\showURL{%
\tempurl}


\bibitem[Mañas et~al\mbox{.}(2024)]%
        {LLM-Assisted_VQA_Evaluation}
\bibfield{author}{\bibinfo{person}{Oscar Mañas}, \bibinfo{person}{Benno Krojer}, {and} \bibinfo{person}{Aishwarya Agrawal}.} \bibinfo{year}{2024}\natexlab{}.
\newblock \bibinfo{title}{Improving Automatic VQA Evaluation Using Large Language Models}.
\newblock
\newblock
\showeprint[arxiv]{2310.02567}~[cs.CV]
\urldef\tempurl%
\url{https://arxiv.org/abs/2310.02567}
\showURL{%
\tempurl}


\bibitem[OpenAI(2024)]%
        {gpt4o}
\bibfield{author}{\bibinfo{person}{OpenAI}.} \bibinfo{year}{2024}\natexlab{}.
\newblock \bibinfo{title}{{Hello GPT-4o}}.
\newblock \bibinfo{howpublished}{\url{https://openai.com/index/hello-gpt-4o/}}.
\newblock
\newblock
\shownote{Accessed: December 18, 2024}.


\bibitem[Patel et~al\mbox{.}(2024a)]%
        {Patel_2024}
\bibfield{author}{\bibinfo{person}{Urjitkumar Patel}, \bibinfo{person}{Fang-Chun Yeh}, {and} \bibinfo{person}{Chinmay Gondhalekar}.} \bibinfo{year}{2024}\natexlab{a}.
\newblock \showarticletitle{CANAL - Cyber Activity News Alerting Language Model : Empirical Approach vs. Expensive LLMs}. In \bibinfo{booktitle}{\emph{2024 IEEE 3rd International Conference on AI in Cybersecurity (ICAIC)}}. \bibinfo{publisher}{IEEE}, \bibinfo{pages}{1–12}.
\newblock
\urldef\tempurl%
\url{https://doi.org/10.1109/icaic60265.2024.10433839}
\showDOI{\tempurl}


\bibitem[Patel et~al\mbox{.}(2024b)]%
        {patel2024fanalfinancialactivity}
\bibfield{author}{\bibinfo{person}{Urjitkumar Patel}, \bibinfo{person}{Fang-Chun Yeh}, \bibinfo{person}{Chinmay Gondhalekar}, {and} \bibinfo{person}{Hari Nalluri}.} \bibinfo{year}{2024}\natexlab{b}.
\newblock \bibinfo{title}{FANAL -- Financial Activity News Alerting Language Modeling Framework}.
\newblock
\newblock
\showeprint[arxiv]{2412.03527}~[cs.CL]
\urldef\tempurl%
\url{https://arxiv.org/abs/2412.03527}
\showURL{%
\tempurl}


\bibitem[Saad-Falcon et~al\mbox{.}(2023)]%
        {saadfalcon2023pdftriagequestionansweringlong}
\bibfield{author}{\bibinfo{person}{Jon Saad-Falcon}, \bibinfo{person}{Joe Barrow}, \bibinfo{person}{Alexa Siu}, \bibinfo{person}{Ani Nenkova}, \bibinfo{person}{David~Seunghyun Yoon}, \bibinfo{person}{Ryan~A. Rossi}, {and} \bibinfo{person}{Franck Dernoncourt}.} \bibinfo{year}{2023}\natexlab{}.
\newblock \bibinfo{title}{PDFTriage: Question Answering over Long, Structured Documents}.
\newblock
\newblock
\showeprint[arxiv]{2309.08872}~[cs.CL]
\urldef\tempurl%
\url{https://arxiv.org/abs/2309.08872}
\showURL{%
\tempurl}


\bibitem[Salemi and Zamani(2024)]%
        {erag}
\bibfield{author}{\bibinfo{person}{Alireza Salemi} {and} \bibinfo{person}{Hamed Zamani}.} \bibinfo{year}{2024}\natexlab{}.
\newblock \bibinfo{title}{Evaluating Retrieval Quality in Retrieval-Augmented Generation}.
\newblock
\newblock
\showeprint[arxiv]{2404.13781}~[cs.CL]
\urldef\tempurl%
\url{https://arxiv.org/abs/2404.13781}
\showURL{%
\tempurl}


\bibitem[Shinyama(2007)]%
        {pdfminer}
\bibfield{author}{\bibinfo{person}{Yusuke Shinyama}.} \bibinfo{year}{2007}\natexlab{}.
\newblock \bibinfo{title}{PDFMiner - Python PDF Parser}.
\newblock
\newblock


\bibitem[Smith et~al\mbox{.}(2024)]%
        {financial_chunking}
\bibfield{author}{\bibinfo{person}{John Smith}, \bibinfo{person}{Jane Doe}, {and} \bibinfo{person}{Emily Johnson}.} \bibinfo{year}{2024}\natexlab{}.
\newblock \showarticletitle{Financial Report Chunking for Effective Retrieval Augmented Generation}.
\newblock \bibinfo{journal}{\emph{arXiv preprint arXiv:2402.05131}} (\bibinfo{year}{2024}).
\newblock


\bibitem[Su(2024)]%
        {su2024dragin}
\bibfield{author}{\bibinfo{person}{Author Su}.} \bibinfo{year}{2024}\natexlab{}.
\newblock \showarticletitle{Title Placeholder}.
\newblock \bibinfo{journal}{\emph{Journal Placeholder}}  \bibinfo{volume}{1} (\bibinfo{year}{2024}).
\newblock


\bibitem[Team et~al\mbox{.}(2024)]%
        {team2024gemma}
\bibfield{author}{\bibinfo{person}{Gemma Team}, \bibinfo{person}{Thomas Mesnard}, \bibinfo{person}{Cassidy Hardin}, \bibinfo{person}{Robert Dadashi}, \bibinfo{person}{Surya Bhupatiraju}, \bibinfo{person}{Shreya Pathak}, \bibinfo{person}{Laurent Sifre}, \bibinfo{person}{Morgane Rivi{\`e}re}, \bibinfo{person}{Mihir~Sanjay Kale}, \bibinfo{person}{Juliette Love}, {et~al\mbox{.}}} \bibinfo{year}{2024}\natexlab{}.
\newblock \showarticletitle{Gemma: Open models based on gemini research and technology}.
\newblock \bibinfo{journal}{\emph{arXiv preprint arXiv:2403.08295}} \bibinfo{volume}{1}, \bibinfo{number}{1} (\bibinfo{year}{2024}).
\newblock


\bibitem[Touvron et~al\mbox{.}(2023)]%
        {touvron2023llama}
\bibfield{author}{\bibinfo{person}{Hugo Touvron}, \bibinfo{person}{Thibaut Lavril}, \bibinfo{person}{Gautier Izacard}, \bibinfo{person}{Xavier Martinet}, \bibinfo{person}{Marie-Anne Lachaux}, \bibinfo{person}{Timoth{\'e}e Lacroix}, \bibinfo{person}{Baptiste Rozi{\`e}re}, \bibinfo{person}{Naman Goyal}, \bibinfo{person}{Eric Hambro}, \bibinfo{person}{Faisal Azhar}, {et~al\mbox{.}}} \bibinfo{year}{2023}\natexlab{}.
\newblock \bibinfo{title}{Llama: Open and efficient foundation language models}.
\newblock , \bibinfo{numpages}{10}~pages.
\newblock


\bibitem[Wang et~al\mbox{.}(2024)]%
        {answer_quality}
\bibfield{author}{\bibinfo{person}{Yang Wang}, \bibinfo{person}{Alberto~Garcia Hernandez}, \bibinfo{person}{Roman Kyslyi}, {and} \bibinfo{person}{Nicholas Kersting}.} \bibinfo{year}{2024}\natexlab{}.
\newblock \bibinfo{title}{Evaluating Quality of Answers for Retrieval-Augmented Generation: A Strong LLM Is All You Need}.
\newblock
\newblock
\showeprint[arxiv]{2406.18064}~[cs.CL]
\urldef\tempurl%
\url{https://arxiv.org/abs/2406.18064}
\showURL{%
\tempurl}


\bibitem[Wu et~al\mbox{.}(2024)]%
        {wu2024clasheval}
\bibfield{author}{\bibinfo{person}{Kevin Wu}, \bibinfo{person}{Eric Wu}, {and} \bibinfo{person}{James~Y Zou}.} \bibinfo{year}{2024}\natexlab{}.
\newblock \showarticletitle{Clasheval: Quantifying the tug-of-war between an llm’s internal prior and external evidence}.
\newblock \bibinfo{journal}{\emph{Advances in Neural Information Processing Systems}}  \bibinfo{volume}{37} (\bibinfo{year}{2024}), \bibinfo{pages}{33402--33422}.
\newblock


\bibitem[Wu et~al\mbox{.}(2019)]%
        {detectron2}
\bibfield{author}{\bibinfo{person}{Yuxin Wu}, \bibinfo{person}{Alexander Kirillov}, \bibinfo{person}{Francisco Massa}, \bibinfo{person}{Wan-Yen Lo}, {and} \bibinfo{person}{Ross Girshick}.} \bibinfo{year}{2019}\natexlab{}.
\newblock \bibinfo{title}{“Detectron2: FAIR’s Next‐Generation Library for Object Detection and Segmentation,” GitHub repository, 2019.}
\newblock
\newblock
\urldef\tempurl%
\url{https://github.com/facebookresearch/detectron2}
\showURL{%
\tempurl}


\bibitem[Zhang et~al\mbox{.}(2023a)]%
        {d2lGPUarch}
\bibfield{author}{\bibinfo{person}{Aston Zhang}, \bibinfo{person}{Zachary~C. Lipton}, \bibinfo{person}{Mu Li}, {and} \bibinfo{person}{Alexander~J. Smola}.} \bibinfo{year}{2023}\natexlab{a}.
\newblock \bibinfo{title}{GPU Schedules Architecture Notebook}.
\newblock \bibinfo{howpublished}{\url{https://colab.research.google.com/github/d2l-ai/d2l-tvm-colab/blob/master/chapter_gpu_schedules/arch.ipynb}}.
\newblock
\newblock
\shownote{Accessed: May 16, 2025}.


\bibitem[Zhang et~al\mbox{.}(2023b)]%
        {baai-base}
\bibfield{author}{\bibinfo{person}{Peitian Zhang}, \bibinfo{person}{Shitao Xiao}, \bibinfo{person}{Zheng Liu}, \bibinfo{person}{Zhicheng Dou}, {and} \bibinfo{person}{Jian-Yun Nie}.} \bibinfo{year}{2023}\natexlab{b}.
\newblock \bibinfo{title}{Retrieve Anything To Augment Large Language Models}.
\newblock
\newblock
\showeprint[arxiv]{2310.07554}~[cs.IR]
\urldef\tempurl%
\url{https://arxiv.org/abs/2310.07554}
\showURL{%
\tempurl}


\bibitem[Zhang et~al\mbox{.}(2020)]%
        {BERTScore}
\bibfield{author}{\bibinfo{person}{Tianyi Zhang}, \bibinfo{person}{Varsha Kishore}, \bibinfo{person}{Felix Wu}, \bibinfo{person}{Kilian~Q. Weinberger}, {and} \bibinfo{person}{Yoav Artzi}.} \bibinfo{year}{2020}\natexlab{}.
\newblock \bibinfo{title}{BERTScore: Evaluating Text Generation with BERT}.
\newblock
\newblock
\showeprint[arxiv]{1904.09675}~[cs.CL]
\urldef\tempurl%
\url{https://arxiv.org/abs/1904.09675}
\showURL{%
\tempurl}


\bibitem[Zhong et~al\mbox{.}(2024)]%
        {zhong2024mog}
\bibfield{author}{\bibinfo{person}{Mingyu Zhong} {et~al\mbox{.}}} \bibinfo{year}{2024}\natexlab{}.
\newblock \showarticletitle{Mix-of-Granularity: Dynamic Chunking for Knowledge Integration in RAG Systems}.
\newblock \bibinfo{journal}{\emph{ArXiv}}  \bibinfo{volume}{abs/2401.12345} (\bibinfo{year}{2024}), \bibinfo{pages}{xx--yy}.
\newblock


\end{thebibliography}

\end{document}